\documentclass[10pt,letterpaper]{article}
\usepackage[top=0.85in,left=2.75in,footskip=0.75in]{geometry}

\usepackage{amsmath,amssymb}

\usepackage{changepage}

\usepackage{textcomp,marvosym}

\usepackage{cite}

\usepackage{nameref,hyperref}

\usepackage[nopatch=eqnum]{microtype}
\DisableLigatures[f]{encoding = *, family = * }

\usepackage[table]{xcolor}

\usepackage{array}

\newcolumntype{+}{!{\vrule width 2pt}}

\newlength\savedwidth

\raggedright
\usepackage[aboveskip=1pt,labelfont=bf,labelsep=period,justification=raggedright,singlelinecheck=off]{caption}

\makeatletter
\renewcommand{\@biblabel}[1]{\quad#1.}
\makeatother

\usepackage{lastpage,fancyhdr,graphicx}
\usepackage{epstopdf}
\fancyheadoffset[L]{2.25in}
\fancyfootoffset[L]{2.25in}
\begin{document}
\vspace*{0.2in}

\begin{flushleft}
{\Large
\textbf{AI application gives users real-time feedback on the level of peace in the social media videos they watch}
}
\newline
\\
Pranav Gilda\textsuperscript{1,2\Yinyang},
Prakhar Dungarwal\textsuperscript{1,2\Yinyang},
Ariel Thongkham\textsuperscript{1,3\Yinyang},
Elena T. Ajayi\textsuperscript{4,2\Yinyang},
Shaifali Choudhary\textsuperscript{1,2\Yinyang},
Teresa Mondria Terol\textsuperscript{5,2\Yinyang},
Christine Lam\textsuperscript{6,2\Yinyang},
Jessica Pardim Araujo\textsuperscript{7\Yinyang},
Maegan McFadyen-Mungali\textsuperscript{7\Yinyang},
Larry S. Liebovitch\textsuperscript{2\Yinyang*},
Peter T. Coleman\textsuperscript{7,2\Yinyang},
Harry West\textsuperscript{8\Yinyang},
Kate Sieck\textsuperscript{9\Yinyang},
Scott Carter \textsuperscript{9\Yinyang}
\\
\bigskip
\textbf{1} Data Science Institute, Columbia University, New York, New York, United States of America
\\
\textbf{2} Advanced Consortium on Cooperation, Conflict, and Complexity, Columbia University, New York, New York, United States of America
\\
\textbf{3} Computer Science, Columbia University, New York, New York, United States of America
\\
\textbf{4} Data Science, St, John's University, New York, New York, United States of America
\\
\textbf{5} Quantitative Methods in the Social Sciences, Columbia University, New York, New York, United States of America
\\
\textbf{6} Barnard College, Columbia University, New York, New York, United States of America
\\
\textbf{7} Teachers College, Columbia University, New York, New York, United States of America
\\
\textbf{8} Department of Industrial Engineering and Operations Research, Columbia University, New York, New York, United States of America
\\
\textbf{9} Harmonious Communities, Toyota Research Institute, Los Altons, California, United States of America
\\
\bigskip

%
%
\Yinyang These authors contributed equally to this work.

* Corresponding author
LSL2140@columbia.edu

\end{flushleft}

\section*{Abstract}
Most people now get their news from videos on social media, such as YouTube and Facebook, rather than through curated journalism. ``We become what we behold.'' The content and tone of language plays an essential role in starting or ending conflicts. ``Hate Speech'' can enhance conflict, ``Peace Speech''  can enhance peace. We developed an application that measures, in real time, these aspects of speech from YouTube videos, which can give users helpful feedback on their own media diet. We used two approaches:\\
1) supervised machine learning. Language in the text of online news media text was tagged by surveys that measure the level of peace in those countries. One fully connected feedforward and 2 convolutional neural networks trained on that data were $\sim 97\%$ accurate in predicting levels of peace in the test set and $\sim 70\%$ accurate in another distinct news text data set, but did not generalize to YouTube videos, suggesting that written text is different than transcribed spoken language.\\
2) social science dimensions. There is no similar external data to tag the text in the YouTube video transcripts. We therefore used 2 word-level sentiment analysis (SA) and 6 context-level large language models (LLMs) to measure 5 social dimensions in peace identified by 59 social science studies: compassion-contempt, news-opinion, promotion-prevention, creativity-order, nuance-simplification. LLMs more closely matched the values by 3 human coders on 52 videos, $r^2\sim0.60$ than SA, at  $r^2\sim0.03$.\\
Results: LLMs successfully measured social dimensions important in peace in YouTube videos, compared to human coders.  These results serve as the basis of an analysis engine that can give users and content creators feedback on their own media diet and creations.

\section*{Author summary}
Most people now get their news from videos on social media,  We developed a computer application that gives people watching YouTube videos real time feedback on the level of peacefulness of their own media diet.  This chrome extension uses AI to measure important aspects of peace in the video transcript, such as the range from compassion to contempt. Making viewers more self-aware of their own media diet could improve their behavior in real life and help content creators better understand the tone of their own media creations.\\

\clearpage
\newgeometry{top=0.85in,left=1in,right=1in,footskip=0.75in}

\section*{Introduction}
The content and tone of the language that people use plays an essential role in starting or ending conflicts.  Peace studies have analyzed ``Hate Speech'' that leads to conflict and violence \cite{Kimotho2016, Soral2018, Ezeibe2021, Ramos2024} But peace is not just the absence of conflict or violence. Recent studies of ``Positive Peace'' have focused on understanding the social processes that reinforce peaceful behaviors \cite{Coleman2020, Fry2006, Deutsch2016, Diehl2016, Goertz2016, Mahmoud2017} including the importance of language, that they called ``Peace Speech'' \cite{GomesdeMatos2000, Friedrich2007, Bolivar2011, Ngabonziza2013, Friedrich2019}.

\textbf{``We become what we behold''} \cite{quote}. 71\% of people aged 16 to 40 years old now get most of their news from videos on social media, such as YouTube and Facebook, rather than through curated journalism \cite{20-40}. These sites and their content creators earn revenue from ads nearby their video content and so prioritize activation through negative emotions to increase screen time and clicks. How can we measure the levels of ``Hate Speech" and ``Peace Speech" on these sites?

Starting about 35 years ago, natural language processing, using machine learning and neural networks, analyzed the language that leads to conflicts \cite{Murphy2024}. These studies identified the variables, such as n-grams (small sets of words) or topics, that were most correlated with established peace indices such as the GPI (Global Peace Index) \cite{Voukelatou2020} or specific events recorded in event data bases such as CAMEO (Conflict and Mediation Event Observations) \cite{GDELT} or ACLED (Armed Conflict Location \& Event Data) \cite{ACLED}. These models also analyzed time series data \cite{Voukelatou2022} and used knowledge graphs ``to connect every person, organization, location, count, theme, news source, and event across the planet into a single massive network that captures what’s happening around the world, showing the interconnections between events and actors" \cite{GDELT2}. 

That work has had some success in the classification, time series analysis, and predictions of ``Hate Speech''. It is only recently, that machine learning methods have been applied to study ``Peace Speech'' \cite{PeterA2023,plosone,Karamolegkou2026}. Much work is now in progress to determine the properties of ``Peace Speech'', illuminate the social processes that underlie it, and suggest novel ways that knowledge can be used to promote peace. In our previous work, we used both conventional and novel methods, to better understand ``Positive Peace" and the nature of `Peace Speech''. Those studies:

\begin{itemize}
\item transformed knowledge graphs (also known as causal loop diagrams and concept maps) created by workshops of social scientists and conflict resolution practitioners, into sets of ordinary nonlinear differential equations, which were then analyzed as dynamical systems to identify attractors of peace and conflict \cite{lsl}.

\item used data-driven rather than hypothesis-driven, supervised machine learning: logistic regression, random forest, support vector machines, decision trees, and embeddings to find the words of highest feature importance that classify lower vs. higher peace countries, which were unexpectedly different from those words hypothesized by social scientists, and clustered those words into topics using k-means, principal component analysis, and large language models \cite{plosone} \cite{Tushar} \cite{Kevin_ML}.

\item organized 6 workshops with journalists, poets, and experts in peace studies, social psychology, anthropology, from Bangladesh, Canada, India, Jamaica, Kenya, Nigeria, Sri Lanka, the United Kingdom, and the United States, that helped us to interpret the data science results described in the previous item  \cite{DougBook}

\item deployed large language models (LLMs) with retrieval augmented prompt generation (RAG) to measure the quantitative levels of social processes, such as positive and negative intergroup reciprocity, in lower and higher peace countries  \cite{Kevin_RAG}.
\end{itemize}

In this paper, we report that we have now developed a chrome extension application, BAIT,  that measures, in real time, aspects of ``Peace Speech'' from YouTube videos, to give users helpful feedback on their own media consumption and to help content creators, journalists, researchers, and platforms better understand the tone of their media creation. BAIT is available for download at: https://bait.fyi/download.

\vspace{\baselineskip}
We used two approaches to develop the back-end analysis engine:
\begin{itemize}
\item supervised machine learning. Language in the text of online news media text was tagged by surveys that measure the level of peace in those countries. One fully connected feedforward and 2 convolutional neural networks were trained on that data. These models trained were 70\% accurate in analyzing other text news online news sources, but were very poor in analyzing the text transcripts of the YouTube videos. This suggests that written text has somewhat different language properties than the transcription of spoken text.
\item social science dimensions. There is no similar external data to tag the text in the YouTube video transcripts. We therefore used 2 word-level sentiment analysis (SA) and 6 context-level large language models (LLMs) to measure 5 social dimensions in peace identified by 59 social science studies: compassion-contempt, news-opinion, promotion-prevention, creativity-order, nuance-simplification. The values of these social dimensions measured by the LLMs more closely matched, $r^2=0.60$, the values by 3 human coders on 52 videos than SA, $r^2=0.03$. Using these LLMs means that BAIT can be a useful tool to rate videos at scale and use those results as the independent variable in subsequent studies on the effects of these videos on behavior. 
\end{itemize}

We used Human Centered Design (HCD) to develop the front-end user interface (UX/UI), Feedback from our team members, students in a design class, and a broader set of people were evaluated to create an iterative set of designs.

\section{Methods: news data}
\subsection*{Strategy}
There are many surveys that measure levels of peace in a country conducted by international organizations, national governmental agencies, non-profit groups, and for-profit consulting companies \cite{plosone}. Thus, we can use that data to tag samples of online news text from different countries with their level of peace to form training and testing sets for supervised learning. We also tested the accuracy of models trained on one dataset analyzing another dataset of different countries, see Fig. 1.


\begin{figure}[h]
\centering
    \includegraphics[width=0.5\textwidth]{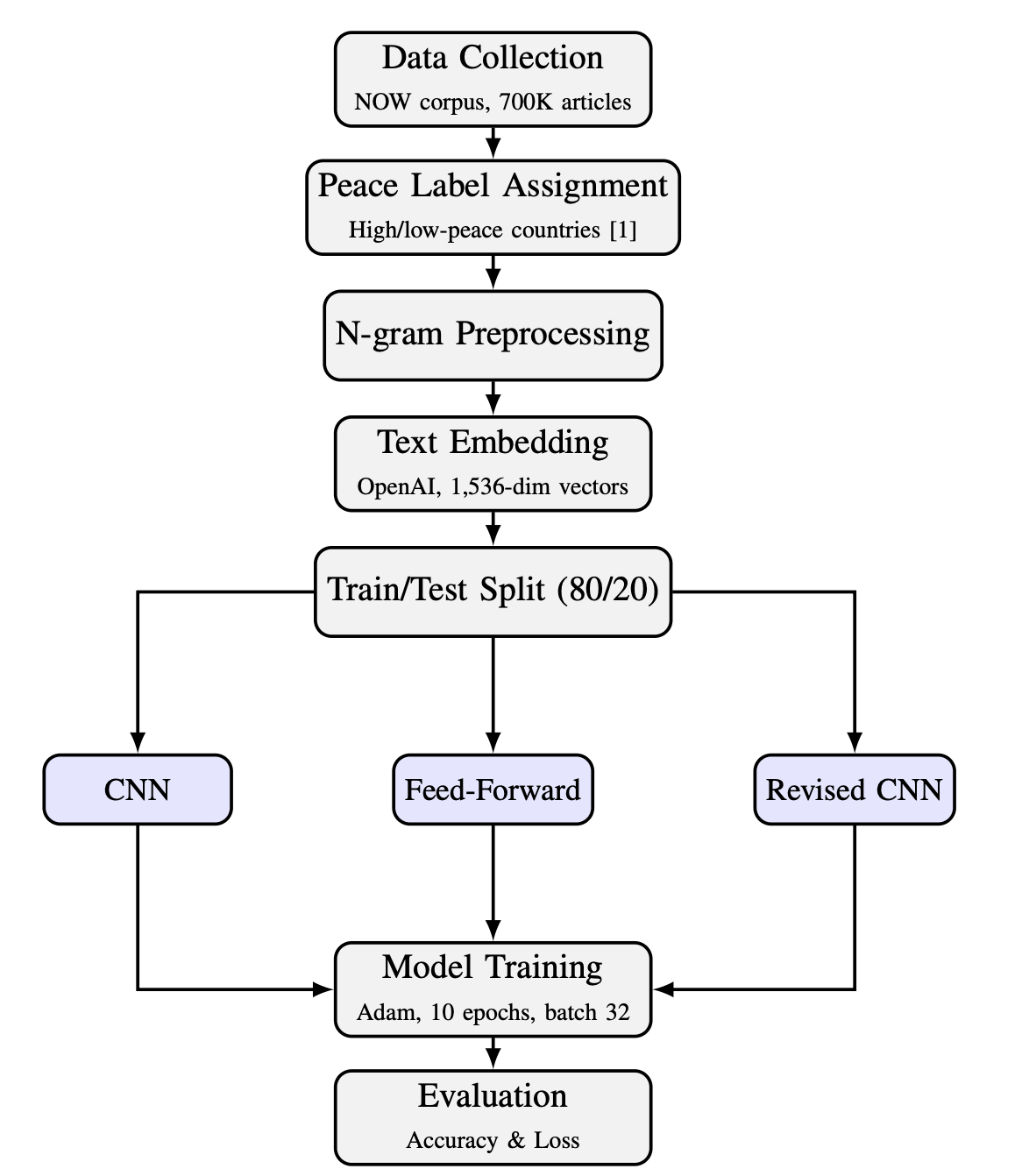}
\caption{Neural network training and evaluation pipeline for peace classification.}
\label{fig:fig1}
\end{figure}

\subsection*{Neural network classification of peace}
\subsubsection*{Data Preparation} \hfill \break
A large-scale training dataset was constructed from the News on the Web (NOW) corpus \cite{NOW}  comprising approximately 700{,}000 news articles from 18 countries. Each article included metadata specifying its country of origin and the associated peace level. High- and low-peace labels were assigned following the method described by Liebovitch et al. \cite{plosone}, which maps linguistic features of national media to established peace indices.

To prepare the data, n-gram preprocessing was applied to capture multi-word contextual patterns and reduce sparsity. The processed text was then embedded using the OpenAI text-embedding-3-small model, which converts each article into a 1{,}536-dimensional vector representation encoding its semantic meaning. Of the 18 countries represented, 10 were identified as high-peace, resulting in a balanced distribution for unbiased model training and evaluation.\\

\subsubsection*{Model Architectures} \hfill \break
Three neural architectures were developed to perform binary classification of articles as high-peace or low-peace:

\begin{itemize}
    \item \textbf{CNN Model:} Two convolutional layers (64 and 32 filters; kernel size = 3, ReLU activation) were followed by a flattening layer and two dense layers (128 and 64 units) with a 0.3 dropout rate for regularization and a sigmoid output. This configuration was designed to extract localized semantic patterns within the embeddings, capturing phrases or n-grams indicative of peaceful discourse.
    \item \textbf{Feed-Forward Model:} A fully connected network of four dense layers (512, 256, 128, and 64 units), each followed by 0.3 dropout, concluded with a sigmoid output layer. This deeper architecture enabled the model to learn global semantic relationships across articles, representing broader conceptual differences in peace-related language.
    \item \textbf{Revised CNN Model:} The revised design incorporated a max-pooling layer (pool size = 2) between convolutional layers, enabling dimensionality reduction and improved generalization. It retained the two fully connected layers (128 and 64 units) from the original CNN, followed by a sigmoid classifier.
\end{itemize}

These architectures were selected to compare the performance of spatially sensitive versus fully connected learning strategies on high-dimensional text embeddings.\\

\subsubsection*{Training procedure} \hfill \break
The dataset was divided into 80\% training and 20\% testing subsets using a fixed random seed to ensure reproducibility. For the convolutional models, each input sample was represented as a sequence of 1{,}536 elements with a single feature channel, while the feed-forward network received the same 1{,}536-dimensional vectors in flattened form. All models were trained for 10 epochs with a batch size of 32, using the Adam optimizer (learning rate = 0.001) and binary cross-entropy loss. Accuracy served as the primary evaluation metric.

Training histories were recorded to monitor loss and accuracy over epochs, assessing both convergence and generalization on the test set. Dropout regularization (0.3) was applied to mitigate overfitting. No separate validation set was used, as the objective was not hyperparameter tuning but rather establishing a consistent baseline for the networks’ ability to classify high-peace versus low-peace texts. All models were trained on GPU-enabled hardware using \texttt{TensorFlow 2.15} and \texttt{Python 3.10}.\\

\section*{Methods: YouTube data}

\subsection*{Strategy}
The news data models were trained by tagging online news text with the levels of peace from external traditional peace indices. There are no similar external peace measures to tag YouTube video transcripts.  Therefore, we based our YouTube models both on: 1) feedback that we received from peace experts and journalists reviewing our previous machine learning results \cite{DougBook} and 2) on social dimensions identified by 59 studies over the last 40 years as important measures of the level of peace, including studies on: integrative complexity, political framing meta-analysis, tightness/looseness, and many others \cite{gelfand,Conway2001,meta} documented in the Supporting Information  \cite{Coleman2026}. These dimensions are:
\begin{center}
compassion --- contempt\\
news --- opinion\\
promotion --- prevention\\
creativity --- order\\
nuance --- simplification\\
\end{center}
Higher peace societies are predicted to evidence more: compassion, news, promotion, creativity, and nuance. Lower peace societies are predicted to evidence more: contempt, opinion, prevention, order, and simplification.

\subsection*{Analytical models for YouTube transcripts}
The failure of the news-trained networks demonstrated that peace in video content is not defined by specific keywords (topics) but by the manner in which those topics are discussed (tone and framing). While Neural Networks excelled at detecting topic-based patterns in news, they failed to capture the conversational volatility of video. We therefore pivoted to models designed to detect emotional tone (GoEmotions) and contextual framing (LLMs) shown in Table \ref{tab:LLMs}.

\begin{table}[htbp]
\centering
\caption{Models Used \cite{Liu2019,Demszky2020,GeminiTeam2023,OpenAI2023,OpenAI2025gpt5}}
\begin{tabular}{l l l}
\hline
\textbf{Abbreviation} & \textbf{Application} & \textbf{Source} \\
\hline
G 2.5 Flash & Gemini 2.5 Flash & Google \\
G 2.5 Flash +R & Gemini 2.5 Flash + RoBERTa & Google \\
G 3 Pro & Gemini 3 Pro Preview & Google \\
GoEmotions & GoEmotions & Google \\
GoEmotions +R & GoEmotions +RoBERTa & Google \\
GPT-4o & ChatGTP-4o & OpenAI \\
GPT-4o +R & ChatGTP-4o + RoBERTa & OpenAI \\
GPT 5.1 & ChatGPT 5.1 & OpenAI \\
Humans & Coding & Experts \\
\hline
\label{tab:LLMs}
\end{tabular}
\end{table}

\subsection*{Google GoEmotions}
Our first approach utilized Google's GoEmotions, a RoBERTa-based transformer model trained on a large corpus of Reddit comments to classify text into 28 emotion categories \cite{Liu2019,Demszky2020}. We hypothesized that we could map these fine-grained emotions to our five broader peace dimensions. However, this approach revealed several significant limitations:

\begin{itemize}
    \item \textbf{Limited Context Window:} The model analyzes text in small, discrete chunks (e.g., sentence-by-sentence), lacking a broader, holistic understanding of the video's full narrative arc. While we aggregated scores at the paragraph level, this post-processing step could not fix the underlying lack of context.

    \item \textbf{High Neutrality Baseline:} The model, trained on a different domain (Reddit comments), frequently classified large portions (40-70\%) of the YouTube transcripts as ``neutral.'' This high baseline score obscured the more subtle emotional cues indicative of bias or nuance, limiting the model's utility.

    \item \textbf{The Averaging-Out Problem:} A critical failure mode of simple aggregation is its inability to capture emotional volatility. For example, a transcript that expressed strong contempt for one topic (e.g., immigrants) and later expressed strong praise for another (e.g., a policy) would have its scores averaged, misleading the results. This process effectively erased the very nuance we intended to measure.

    \item \textbf{Deterministic weighted mapping:}
    We attempted a heuristic approach by mapping GoEmotions outputs to a scalar valence score (e.g., \textit{joy/admiration} = +1.0, \textit{anger/disgust} = -1.0, \textit{neutral} = 0.0). However, this deterministic mapping failed to correlate with the dimensions of the peace because it ignored the context.
\end{itemize}

These limitations made it clear that while the GoEmotions scores were valuable as a quantitative \emph{feature}, they were insufficient as a standalone \emph{analyzer} for our complex peace dimensions.

\textbf{Motivation for a synthesis model:}
This led us to hypothesize that a more robust solution would be a multi-stage process. We needed a model that could not only see the quantitative emotional data from GoEmotions but also interpret it within the full context of the entire transcript.

This motivated our pivot to Large Language Models (LLMs) as a synthesis tool. We theorized that LLMs, while potentially subjective when used in isolation, could be ``grounded'' by providing the GoEmotions scores as a structured, quantitative input. This dual-input approach, giving the LLM both the text and its emotional profile—was designed to improve trustworthiness, observability, and provide a tighter feedback loop for our team to evaluate the model's reasoning.

\subsection*{Large language models (LLMs)}
In parallel with the GoEmotions analysis, we explored the capabilities of Large Language Models (LLMs) to score transcripts along the same peace dimensions. Our methodology evolved through several stages of prompt engineering, moving from a general-purpose prompt to a final, sophisticated model that synthesized textual data with quantitative emotional analysis.

\paragraph{Initial Prompting (news media framework)}
We initially adapted a comprehensive prompt designed for a related project analyzing formal news articles analyzing positive and negative intergroup reciprocity \cite{Kevin_RAG}. This prompt, tested on models such as OpenAI's GPT-3 and GPT-4-mini, was highly detailed. It provided a role for the AI (``media analyst''), a +5 to -5 scoring scale, and multiple, detailed examples for each of the five dimensions.

While effective for structured news text, this prompt proved \emph{overly complex and ill-suited} for the conversational, faster-paced, and stylistically different nature of YouTube video transcripts. The extensive rules and examples did not generalize well to the broader domains, leading to inconsistent and unreliable scoring.

\paragraph{Iterative refinement (simplistic-generalizable prompt)}
Recognizing the limitations of the initial prompt, we engaged in an iterative engineering process to develop a version optimized for YouTube transcripts. The primary goals were to simplify the instructions leading to better generalizability, maintain the core 5-dimensional analysis.

\subsection*{Human feedback}
To establish a rigorous benchmark for evaluating our automated approaches, we collaborated with human experts specializing in peace research and conflict studies. Our team of peace researchers, including social psychologists, anthropologists, and conflict resolution specialists from Columbia University's Morton Deutsch International Center for Cooperation and Conflict Resolution, independently evaluated a gold standard set of 52 YouTube videos. (Links to the rubrics used and the coders results are in the Supplementary Information section.)

Each expert scored videos across all five dimensions using the same 1-5 scale employed by our computational models. Videos were selected to represent a diverse range of content types, including political commentary, news analysis, social issues discussion, and cultural critique.

Inter-rater reliability analysis validated the robustness of this gold standard, yielding pairwise correlations exceeding $r^2=0.86$ for three dimensions (compassion-contempt, news-opinion, creativity-order) and 100\% agreement within a single-point margin. This high consensus confirms that our peace dimensions are observable and measurable constructs, distinct from random subjective noise.
We aggregated human scores by averaging across all available raters for each video-dimension combination, with sample sizes ranging from N=32 to N=47 depending on missing data patterns. The resulting gold standard dataset provides mean scores, standard deviations, and distributional statistics that serve as ground truth for evaluating model performance, as shown in Table \ref{tab:human_metrics}.

\begin{table}[h]
\centering
\caption{Human Expert Annotation Statistics}
\label{tab:human_metrics}
\begin{tabular}{lcccccc}
\hline
\textbf{Dimension} & \textbf{N} & \textbf{Mean} & \textbf{SD} & \textbf{Min} & \textbf{Max} & \textbf{Median} \\
\hline
compassion - contempt & 47 & 2.89 & 0.94 & 1.0 & 5.0 & 3.0 \\
news - opinion & 34 & 3.06 & 1.43 & 1.0 & 5.0 & 3.4 \\
promotion - prevention & 35 & 2.64 & 1.32 & 1.0 & 5.0 & 2.5 \\
creativity - order & 32 & 2.81 & 1.38 & 1.0 & 5.0 & 3.0 \\
nuance - simplification & 37 & 2.76 & 1.06 & 1.0 & 4.83 & 3.0 \\
\hline
\end{tabular}
\end{table}

\section*{Results: news data}

\subsection*{Evaluation of neural networks for peace classification}
\subsubsection*{Model performance on test data} \hfill \break
Across the three network architectures, all models achieved high performance on the held-out test set from the NOW dataset, as shown in Table \ref{tab:nn_performance}. The convolutional neural network (CNN) achieved a test accuracy of 97.24\%, the fully connected feed-forward network reached 97.48\%, and the revised CNN achieved 96.99\%. These results indicate that the models effectively learned to distinguish linguistic patterns associated with high- and low-peace discourse within structured news text.

\begin{table}[htbp]
\caption{Neural Network Performance Across Datasets}
\centering
\small
\begin{tabular}{lccc}
\hline
\textbf{Model} & \textbf{NOW (\%)} & \textbf{2022 Capstone (\%)} \\
\hline
CNN & 97.24 & 72.81 \\
Feed-Forward & 97.48 & 72.47 \\
Revised CNN & 96.99 & 69.91 \\
\hline
\end{tabular}
\label{tab:nn_performance}
\end{table}

Although all three models performed comparably, the feed-forward network’s slightly higher accuracy aligns with its architectural advantage: its dense layers integrate global semantic relationships across all embedding dimensions. This capability is particularly beneficial in a binary classification task such as peace prediction, where subtle distinctions often depend on distributed semantic cues across the entire embedding space. By contrast, the CNNs—while effective at detecting local n-gram-like patterns—may overlook long-range dependencies essential for capturing the overall semantic and emotional tone of written language. Nonetheless, the CNNs’ strong performance demonstrates that both local and global text features contributed meaningfully to the classification of peace-related language. \\

\subsubsection*{Cross data generalization}
To assess the generalizability of the trained models, they were tested on an independent corpus—the 2022 Capstone Peace Speech dataset comprising 600{,}000 news articles from 16 countries (eight high-peace and eight low-peace) \cite{Chen2022}. Each article was labeled according to established peace indices such as the Global Peace Index and the Positive Peace Index. Despite differences in collection methodology and source distribution, the models show substantial cross-dataset transfer in the per-article predictions: 72.81\% for the CNN, 72.47\% for the feed-forward network, and 69.91\% for the revised CNN. When grouped and averaged at the country level, all three networks successfully classified every country in accordance with its peace label. This result suggests that the linguistic features learned from the NOW dataset capture generalizable distinctions in structure, tone, and framing that align with broader indicators of national peacefulness.

\subsubsection*{Comparison between network architectures} \hfill \break
The per-article Capstone evaluation shows that the original CNN transfers slightly better than the revised CNN and performs comparably to the feed-forward network. A likely reason is that the original convolutional design, without the added pooling layer, retains fine-grained local patterns—short semantic or tonal cues that may remain stable across domains. The feed-forward network, meanwhile, leverages global relationships across all embedding dimensions. Overall, the strong performance across models indicates that peace-related linguistic markers exist at multiple levels of representation and can be effectively captured by both convolutional and fully connected architectures. \\

\subsubsection*{Cross-domain generalization to YouTube transcripts} \hfill \break
To explore whether the trained models could generalize beyond formal written text, they were applied to a smaller dataset of 22 YouTube transcripts drawn from five major news organizations: \textit{The New York Times}, \textit{CNN}, \textit{NBC}, \textit{The Washington Post}, and \textit{Breitbart News}. However, when evaluated on this dataset, all networks exhibited severe overgeneralization, classifying 95–100\% of the videos as “high-peace.” This outcome suggests a transfer failure from written to spoken media.

That result may imply that written text from online news sources is different from the transcribed text of spoken online news sources. The linguistic characteristics of video journalism differ substantially from print media—not only in syntax and register but also in the use of immediacy, narrative framing, and emotional activation. Video transcripts often include conversational fillers, hesitations, and speaker attributions that are absent from formal writing. These stylistic and pragmatic differences likely disrupted the networks’ ability to apply the same semantic cues used to identify peace-related language in articles. This finding motivated our subsequent use of emotion-based and large language model (LLM) techniques to better capture nuanced indicators of peace in multimodal and social media contexts. \\

\section*{Results: YouTube data}

The AI models that we used are shown in Table \ref{tab:LLMs}. We compared the predicted values from those models analysis of the transcripts of YouTube videos with those reported by our human coders on five social dimensions. 

\begin{figure}[h]
\centering
\includegraphics[width=0.8\textwidth]{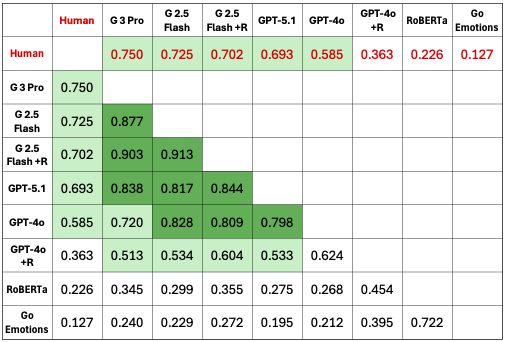}
\caption{Pearson correlation coefficient $r$ between the human coders and predictions of the AI models on the dimension: compassion---contempt. We used the models: Gemini 3 Pro Preview, Gemini 2.5 Flash, GPT-5.1, GPT-4o, RoBERTa, and GoEmotions. +R indicates hybrid models with RoBERTa.}
\label{fig:corr}
\end{figure}

\textbf{Summary of YouTube results} Fig. \ref{fig:corr}, shows the Pearson correlation coefficients between each of the models and the human coders for the values of compassion---contempt. (We followed the convention in these correlation ``heat maps'' using the Pearson correlation coefficient, $r$, which can be either negative or positive, to indicate negative or positive correlations between variables. To indicate the statistical strength of these correlations in the text, we used the appropriate statistical measure, $r^2$, which is the explained variance divided by the total variance.) The sentiment models, such as Google's GoEmotions, based on the valence of individual words, even if enhanced by the transformer RoBERTa, were only weakly correlated, $r ^2\approx 0.03$, with those of our human coders. As we gave that social dimension the highest priority in our first designs of BAIT, we then explored the use of LLMs.

\begin{figure}[h]
\centering
\includegraphics[width=0.75\textwidth]{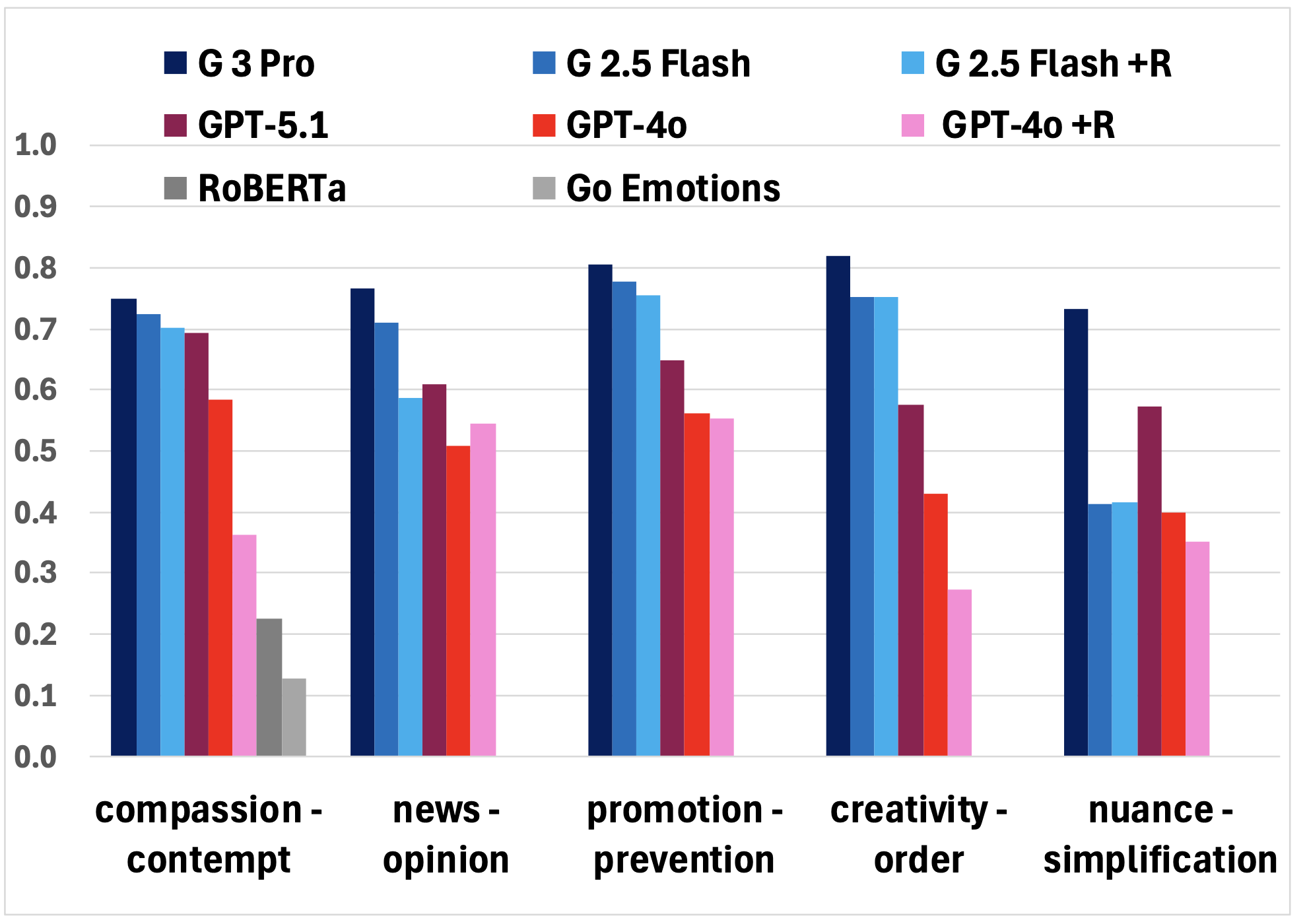}
\caption{Pearson correlation coefficient $r$ between the human coders and predictions of the AI models for all 5 social science dimensions.}
\label{fig:bar}
\end{figure}

As shown in Fig. \ref{fig:bar}, modern large language models can reliably evaluate multidimensional aspects of peace journalism in online video content, achieving correlations, up to $r^2=0.60$, comparable to human inter-rater agreement. This suggests that AI systems can meaningfully scale and augment human media analysis. Gemini 3 Pro Preview achieved the highest correlations across all five dimensions, marking an important improvement over previous iterations in the challenging nuance-simplification dimension. This suggests that the latest reasoning models are beginning to bridge the gap in understanding context-heavy "gray areas" where earlier models struggled.


\textbf{LLMs understand context beyond surface emotion signals.}
Across all dimensions, LLMs substantially outperformed simpler emotion-only baselines, indicating that peace assessment requires contextual reasoning---tone, framing, rhetorical structure, and implicit messaging---rather than raw sentiment. The relatively weaker performance on the nuance---simplistic dimension reflects a fundamental challenge: recognizing multiple perspectives may require deeper external knowledge, pointing toward the value of retrieval-augmented or fact-aware models.

\textbf{Model Architecture and peace-metric alignment.}
This consistent performance advantage suggests that the Gemini architecture may be better aligned with the nuances of sociological scoring tasks, underscoring the importance of empirical evaluation over vendor expectations on domain-specific tasks rather than relying on general benchmarks. Emotion-augmented prompting yielded mixed effects, suggesting that additional signals do not automatically improve reasoning.

\textbf{Multimodality and agents}
Limitations such as dataset size, lack of multimodal cues, and absence of causal evidence highlight clear next steps: integrating audio-visual features, exploring multi-agent assessment pipelines, developing smaller fine-tuned models for edge deployment, and extending across platforms such as TikTok and Twitter.

\section*{Discussion}
There are now many applications that evaluate text, pictures, or video on social media on dimensions of true - false \cite{invid2017, checkmate2024, bbcverify2023}, liberal bias - conservative bias \cite{allsides2012, adfontesmedia2018, groundnews2018}, and human generated - AI generated \cite{oorloff2024, sensityai2019, sightengine}. Recently, RageCheck which ``measures the kind of language designed to provoke emotional reactions rather than inform'' \cite{RageCheck} addresses both tone and content. However, our work is different and unique from those applications cited above as we have used the extensive research findings on ``Hate Speech'' and ``Peace Speech'' to inform what we measure and the machine learning and AI methods that we use to measure it. Unlike much of those other applications, we are also publicly fully documenting our analysis engine and our human testing results.

Our next steps to improve BAIT are to: 1) provide users information on how the peace levels of the videos they watch changes over time, 2) use more extensive human feedback to compare the accuracy of different computational methods to measure the social dimensions in the videos, and 3) expand our testing of the UX/UI interface across a broader group of people.

Our motivation for this project was that making users more self-aware of the level of peace in their own media diet could lead them to more positive social behaviors. The first step needed to test that hypothesis is to be able to measure 'Hate Speech'' and ``Peace Speech'', which is what we have presented here. We now look forward to using that tool to do behavioral testing to measure the effects of BAIT on the choice of videos that users watch and if those choices lead to improved behaviors with others in real life.

Perhaps, this could also lead to a virtuous cycle. If the self-awareness of viewers leads them to watch more videos of higher peace values, then content creators will more likely make videos with higher peace values, and platforms seeking higher user engagement will display more of those videos.

\section*{Conclusions}
``Hate Speech'' can lead to conflict and violence. ``Peace Speech'' can sustain the social processes that reinforce peaceful behaviors. We developed a Chrome extension, BAIT \cite{BAIT}, that accurately measures the levels of social dimensions in videos that are important in peace, as identified by human experts and hypothesis-driven social science studies. The values of the social dimensions in the transcripts of YouTube videos, measured by BAIT using LLMs, correlate $r^2\sim0.60$, with those recorded by human coders.

We also found that neural networks trained on online news text that successfully predicted the levels of peace in other online news text data sets, were not successful in predicting levels of peace in the transcripts of YouTube videos. This may imply that the language properties of online written text differs from that of the transcripts of spoken text, so that each of these may require different analysis tools.

\begin{figure}[htbp]
    \centering
    \includegraphics[width=0.4\textwidth]{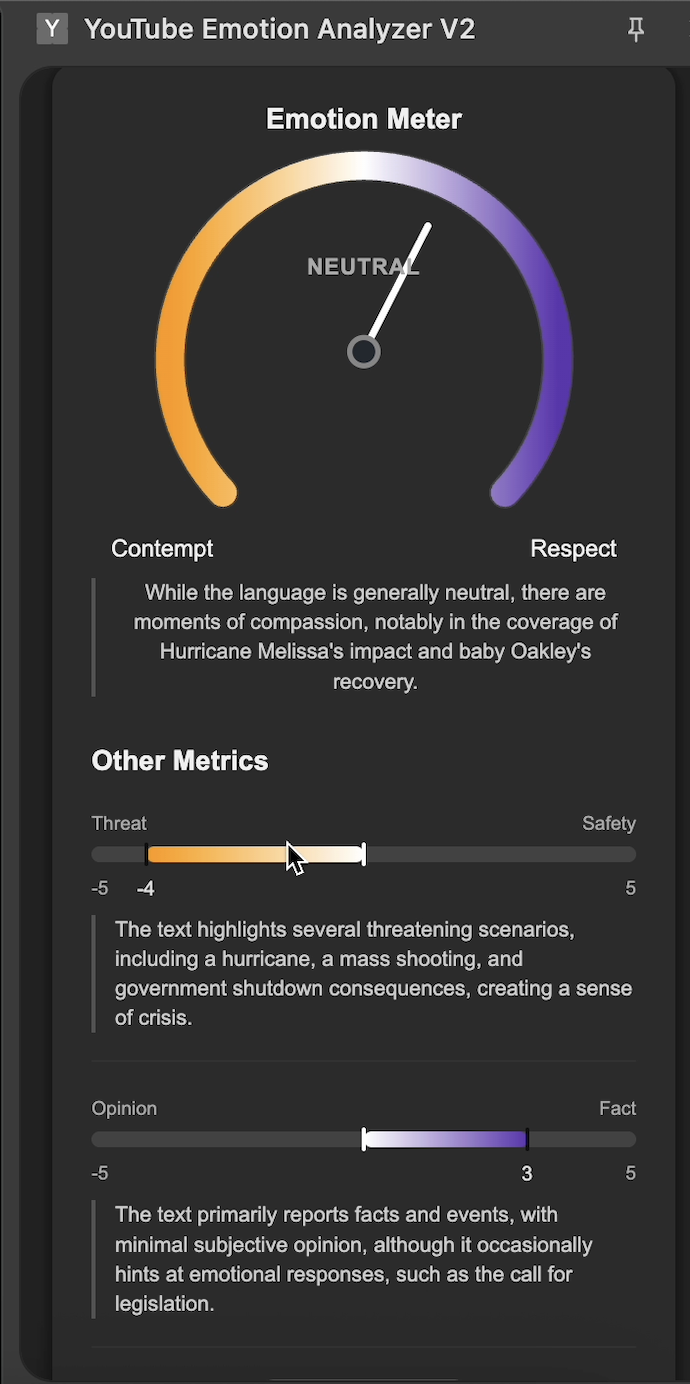}
    \caption{Prototype of the BAIT user interface.}
    \label{fig4}
\end{figure}

The user interface of BAIT shown in Fig. \ref{fig4} sits within the chrome browser next to the video being watched on YouTube. BAIT can provide users useful feedback to increase their awareness of how they are influenced by their own media diet. It can also provide creators feedback on the tone of their media creations. Perhaps, such self-awareness may be a helpful tool in turning down the destructive, polarizing, heat in social media communications.


\section*{Supporting information}
Computer programs\\
analysis engine  \url{https://github.com/pranav-gilda/yt_emotion_dashboard}\\
BAIT chrome extension  \url{https://github.com/bait-extension/extension}\\
\vskip\baselineskip
\noindent Evidence for social dimensions important in peace\\
\texttt{Evidence-Bases.pdf}
\vskip\baselineskip
\noindent Human feedback data\\
coding rubric \texttt{rubric.pdf}\\
coding results \texttt{coded.xlsx}\\
video transcripts \texttt{transcripts.pdf}\\


\end{document}